\pdfoutput=1
\documentclass[10pt]{article}
\usepackage[letterpaper]{geometry}
\usepackage{hicss51}
\usepackage{times}
\usepackage[none]{hyphenat}
\usepackage{url}
\usepackage{latexsym}
\usepackage{indentfirst}
\usepackage{graphicx}
\graphicspath{{images/}}

\PassOptionsToPackage{hyphens}{url}
\usepackage{url}
\usepackage[breaklinks]{hyperref}

\usepackage{cleveref}
\usepackage{multirow}
\usepackage{todonotes}
\usepackage{booktabs}
\usepackage{color}
\usepackage{multirow}
\usepackage{multicol}

\usepackage[square,numbers,sort]{natbib}
\usepackage{xcolor}
\usepackage{adjustbox}

\usepackage{comment}

\title{Utilizing Active Machine Learning for Quality Assurance: A Case Study of Virtual Car Renderings in the Automotive Industry}

\author{Patrick Hemmer\\
Karlsruhe Institute of Technology\\
\underline{patrick.hemmer@kit.edu}\And

Niklas Kühl\\
Karlsruhe Institute of Technology\\
\underline{niklas.kuehl@kit.edu}\And

Jakob Schöffer\\
Karlsruhe Institute of Technology\\
\underline{jakob.schoeffer@kit.edu}}

\date{}

\begin{document}\sloppy
\maketitle
\begin{abstract}
Computer-generated imagery of car models has become an indispensable part of car manufacturers' advertisement concepts. They are for instance used in car configurators to offer customers the possibility to configure their car online according to their personal preferences. However, human-led quality assurance faces the challenge to keep up with high-volume visual inspections due to the car models’ increasing complexity. Even though the application of machine learning to many visual inspection tasks has demonstrated great success, its need for large labeled data sets remains a central barrier to using such systems in practice. In this paper, we propose an active machine learning-based quality assurance system that requires significantly fewer labeled instances to identify defective virtual car renderings without compromising performance. By employing our system at a German automotive manufacturer, start-up difficulties can be overcome, the inspection process efficiency can be increased, and thus economic advantages can be realized.

\end{abstract}

\section{Introduction}
\label{sec:introduction}
Computer-generated imagery (CGI) has become a central element of leading car manufacturers' advertisement and sales concepts.
Virtual car renderings are utilized for various marketing purposes as they enable advertisement campaigns that are efficiently customizable to different markets while not requiring costly physical car prototypes for elaborate photo shoots.
Moreover, CGI introduces to customers the possibility to configure a selected car model in real-time online configurators according to their personal preferences.

Similar to the physical car assembly, high-quality standards are a central requirement for the long-term success of all advertisement offerings. Thus, it is crucial to avoid erroneous content creation.
However, the continuously growing number of design options of all existing car variants including their derivatives poses increasing challenges for currently human-led quality assurance (QA), turning high-volume visual inspection processes into procedural bottlenecks.
Moreover, multiple studies have indicated that the accuracy of human visual inspections may decline with endlessly repetitive routine jobs \cite{chin1982automated,schoonard1973field,see2017role}.

In this context, the continuously evolving capabilities of machine learning (ML) algorithms have led to the successful entry of this technology in many industrial sectors over the last years.
Particularly, in the area of QA, artificial perception systems offer the possibility to support humans through automatic visual inspection \cite{park2016machine}.
Applications range from 
industrial tool wear analysis \cite{treiss2020uncertainty} over printing quality control \cite{villalba2019deep} to quality estimation of porcelain \cite{onita2018quality}.
However, one of the major drawbacks of supervised ML is its dependency on a vast amount of high-quality labeled training data, constituting a major impediment for the launch of such systems in practice \cite{baier2019challenges}.
Even though the amount of available data has been constantly increasing \cite{chen2014big}, labeling of data instances is costly because it often requires the knowledge of domain experts.

To overcome these difficulties for the deployment and operation of ML systems in practice, methods to decrease the data labeling effort, particularly when labeling resources are limited, can be applied, e.g., transfer learning \cite{survey_transfer_learning}, multi-task learning \cite{multi-task-learning}, few-shot learning \cite{snell2017prototypical}, meta-learning \cite{Sun_2019_CVPR} or active (machine) learning (AL) \cite{settles2009active}.
%
The key idea of AL is that a model can accomplish a certain performance level requiring fewer training instances if the most informative instances with regard to its learning process are selected.
Initially, a model is trained on a small labeled data set.
In each AL round, new data instances are chosen by an acquisition function and labeled by an expert. Subsequently, they are added to the labeled data pool. This procedure is repeated until a specific termination criterion is met \cite{settles2009active}. 

In this work, we propose an \emph{active machine learning-based quality assurance} (ALQA) system to assist the manual QA process for the CGI content production of virtual car renderings at Audi Business Innovation GmbH, a subsidiary of AUDI AG.  
Our ALQA system requires significantly less training data and therefore contributes to overcoming the initial start-up barrier of labeling a large number of data instances.
Specifically, our system is able to achieve an average performance (measured in terms of the $F_2$ metric) of $95.81\%$, while reducing the labeling overhead by $35\%$ on average. Identified defective configurations are forwarded by the system to the responsible QA managers for further root cause analysis. 

With our research, we contribute to the body of knowledge with the following five core aspects: First, we propose a novel artifact---an ALQA system---to automatically identify defects in CGI of virtual car models as a support tool for human-led QA.
Second, we evaluate its performance and demonstrate its ability to significantly reduce the manual labeling effort of image data required to train the system while maintaining its prediction accuracy.
Third, we discuss possible economic implications of the system and show its ability to overcome start-up difficulties while increasing the QA efficiency.   
Fourth, by illustrating the feasibility of the artifact, we are the first to introduce deep learning into the QA of CGI within an industrial context---as similar approaches have (so far) only been utilized in the area of video game streaming \cite{zadtootaghaj2020quality}.
Finally, we provide generalizable prescriptive design knowledge by providing precise design principles to address the design requirements of ALQA systems. These ``blueprints'' can serve as a foundation for the design of future artifacts within different domains.

We utilize design science research (DSR) as an overall research design and base our approach on \citet{hevner2010}. They recommend that a DSR project should consist of at least three cycles of investigation: a relevance cycle (addressing the practical problem, see \Cref{sec:relevancecycle}), a rigor cycle (outlining the existing knowledge base, see \Cref{sec:rigorcycle}), and one or multiple design cycles (instantiating and evaluating the research artifact, see \Cref{sec:artifactdesign,sec:artifactevaluation}). We finish our work with a discussion of the potential economic impact of our artifact (\Cref{sec:discussion}) as well as a conclusion (\Cref{sec:conclusion}).

\section{Relevance Cycle: Defining the Problem}
\label{sec:relevancecycle}
 \begin{figure*}[htbp] 
  \centering
  \begin{minipage}[b]{0.24\textwidth}
    \includegraphics[width=\textwidth]{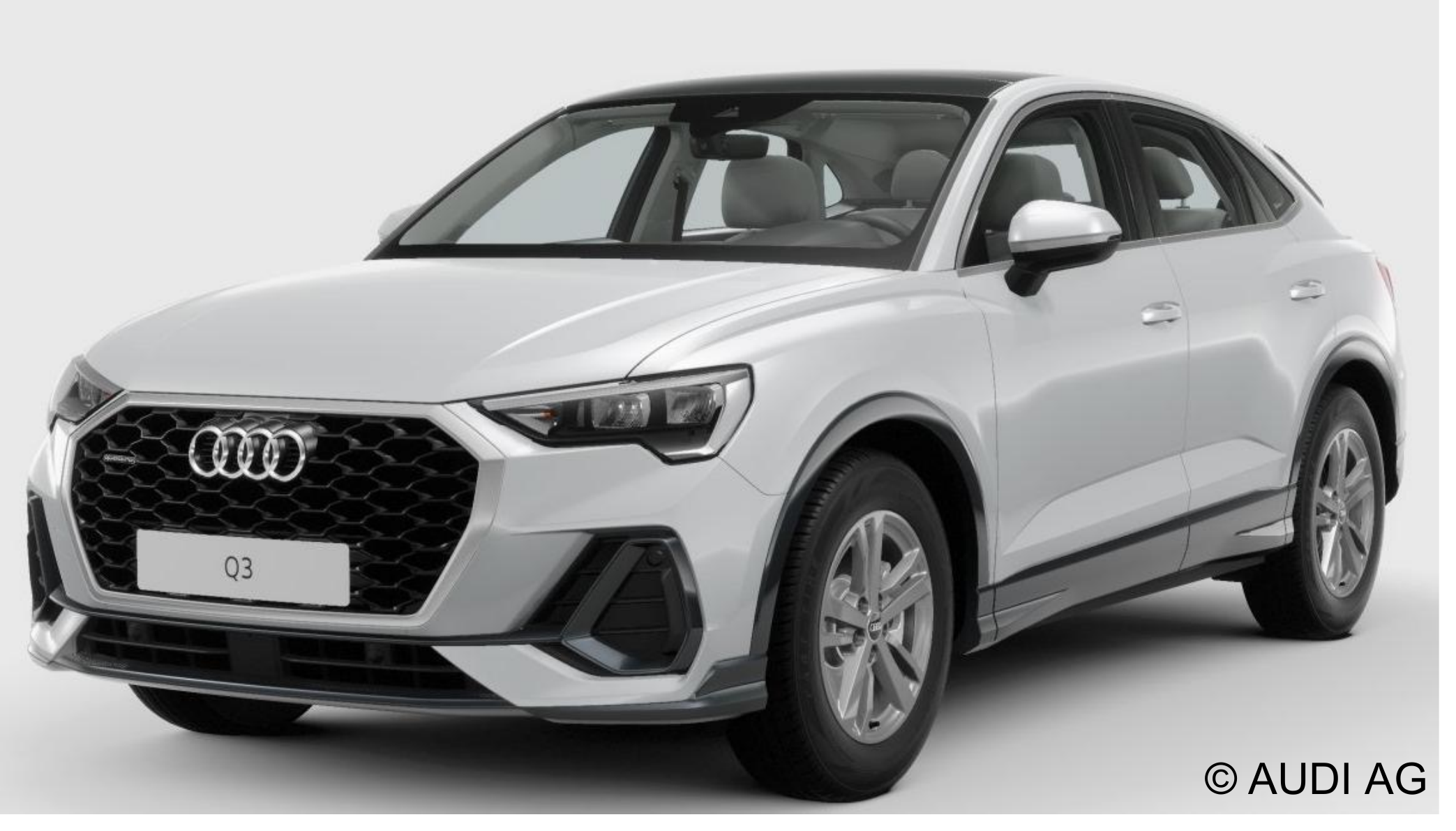}
    \includegraphics[width=\textwidth]{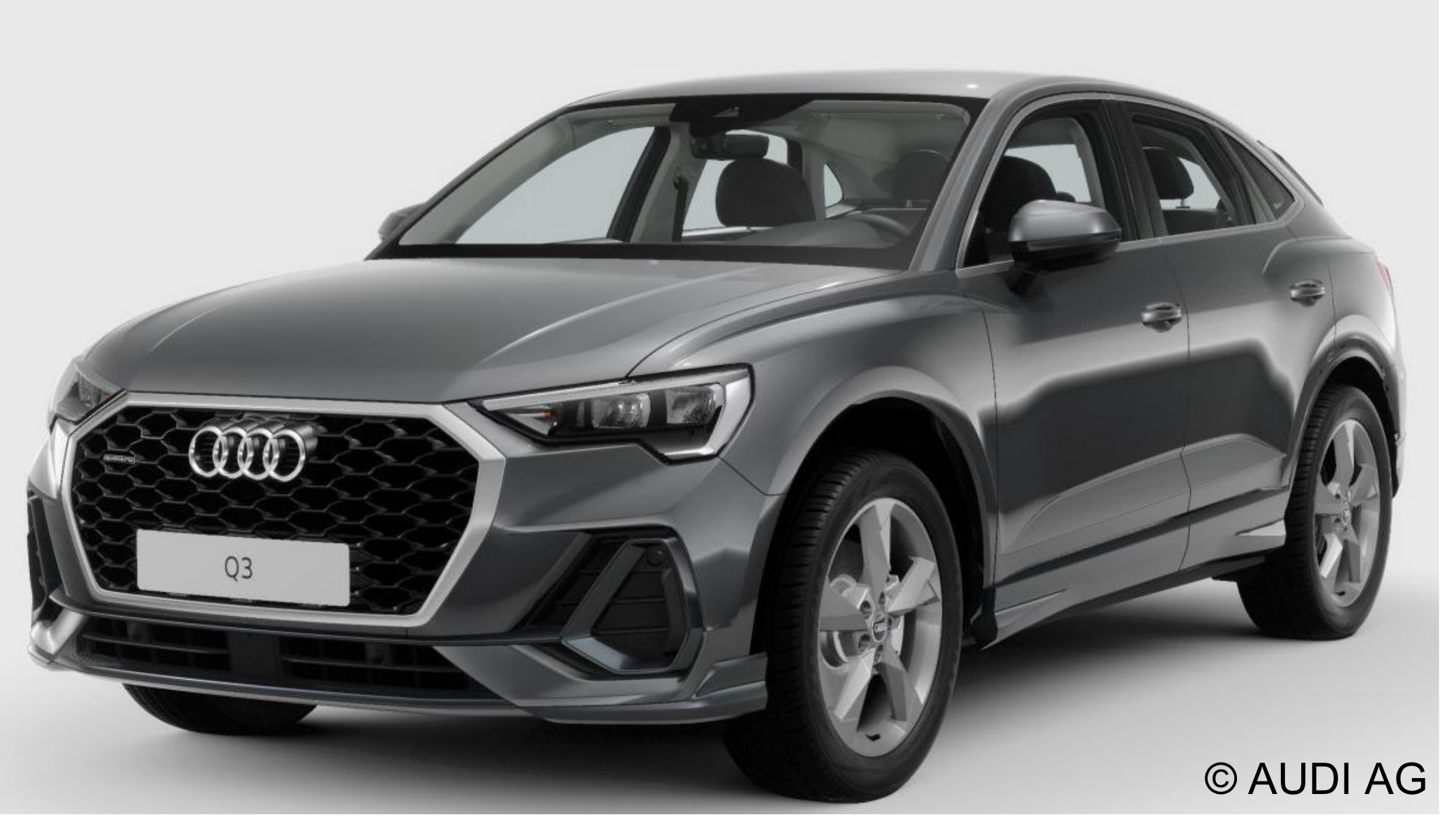}
  \end{minipage}
  \begin{minipage}[b]{0.24\textwidth}
    \includegraphics[width=\textwidth]{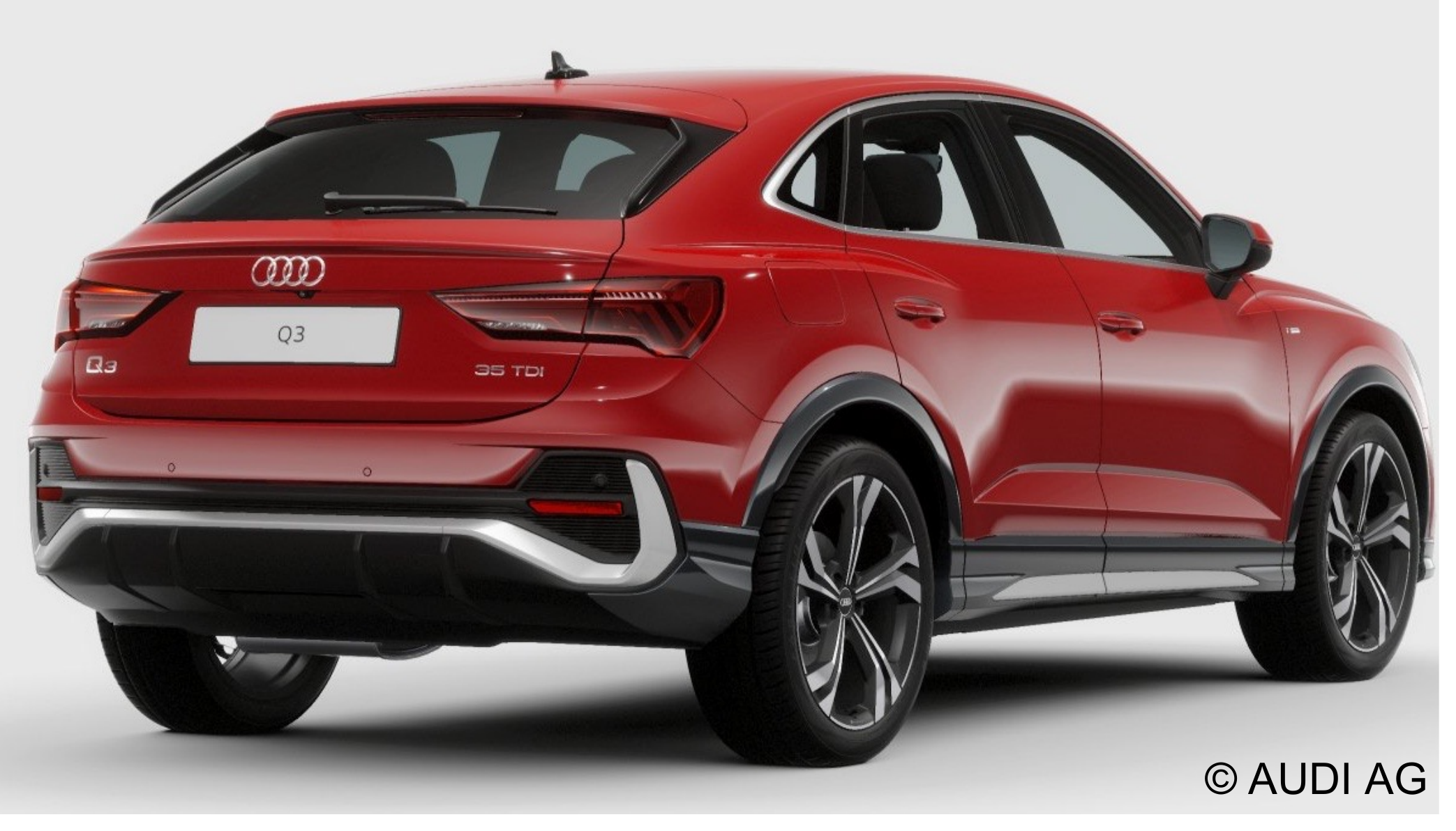}
    \includegraphics[width=\textwidth]{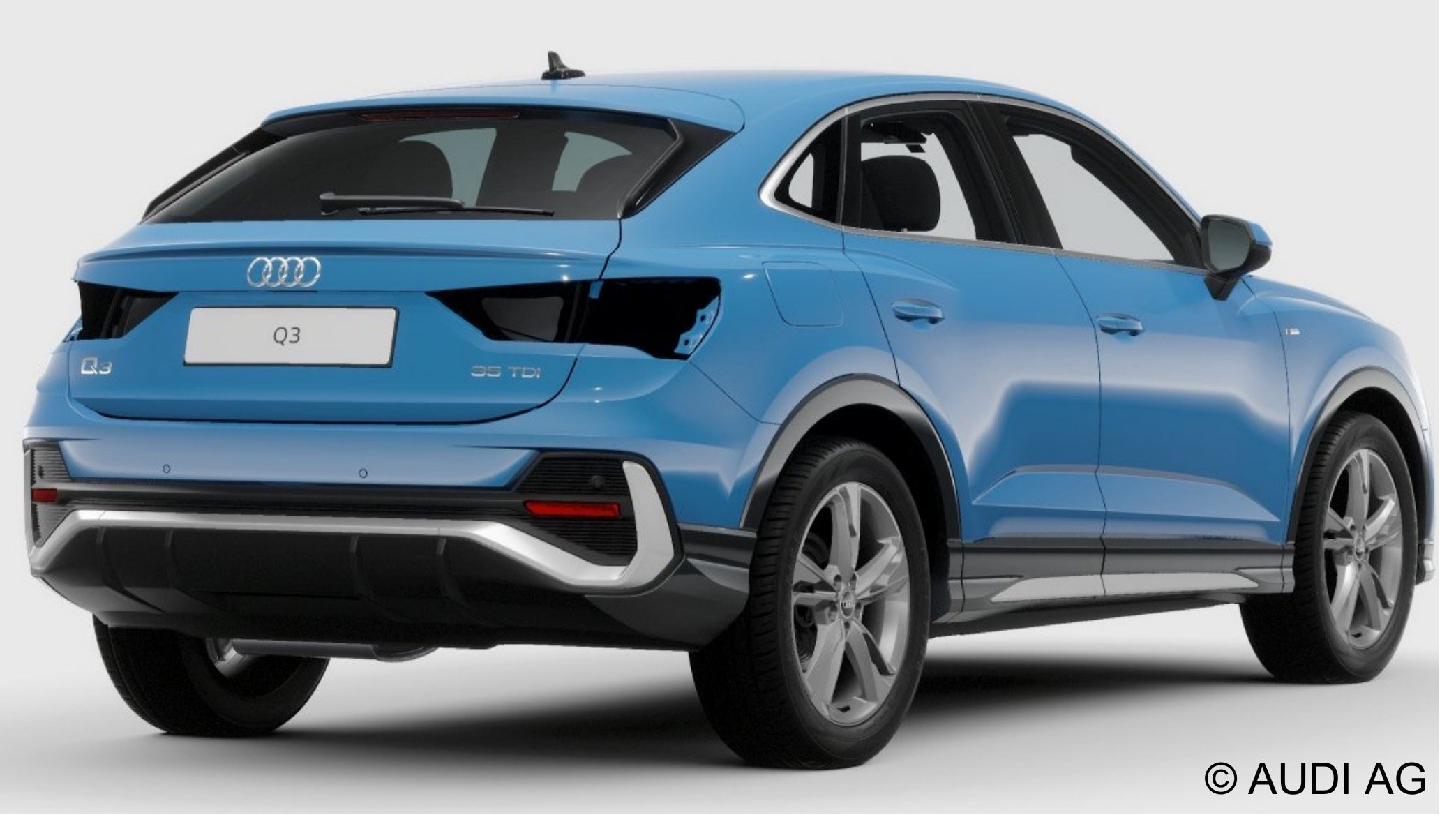}
  \end{minipage}
    \begin{minipage}[b]{0.24\textwidth}
    \includegraphics[width=\textwidth]{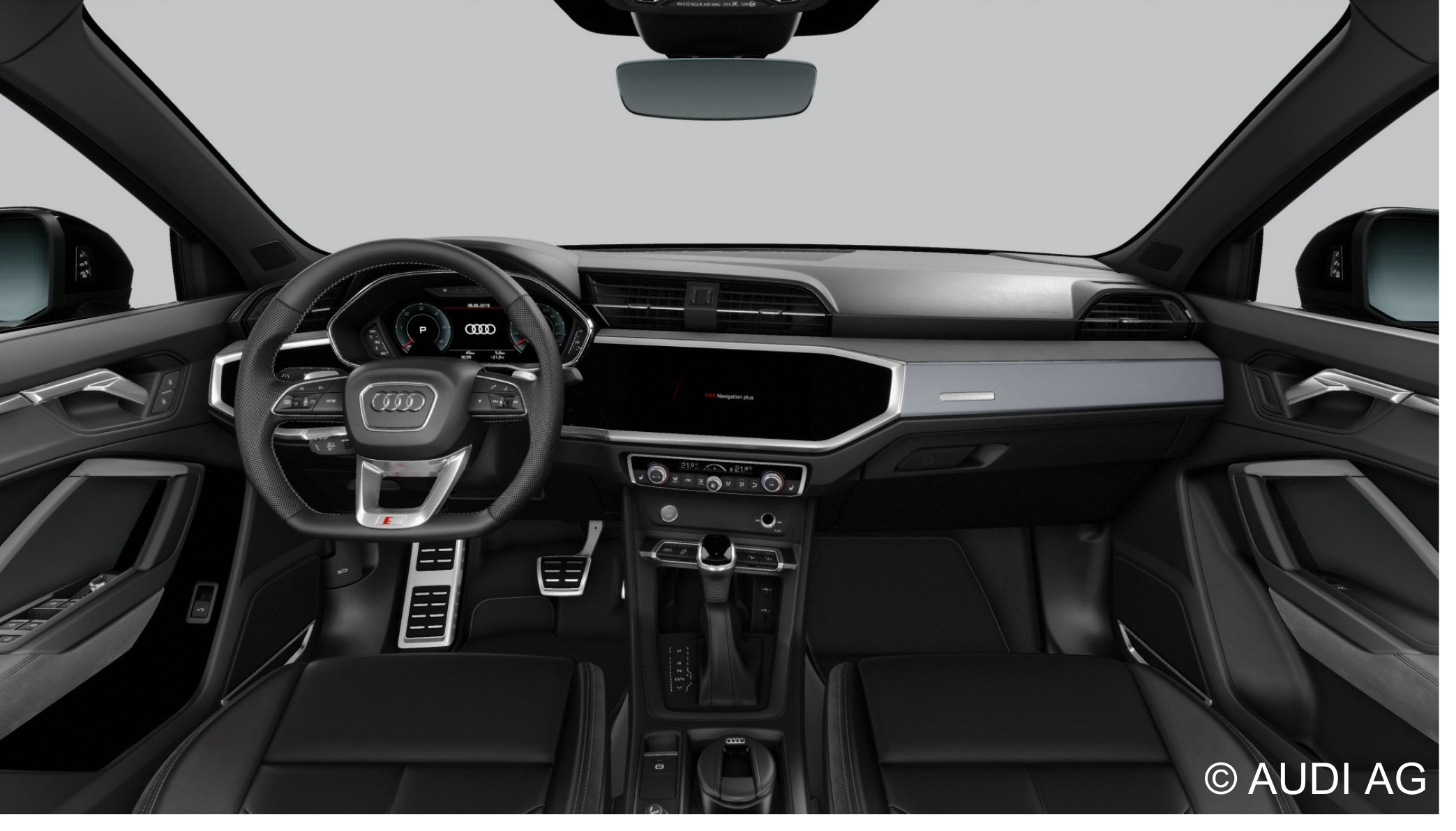}
    \includegraphics[width=\textwidth]{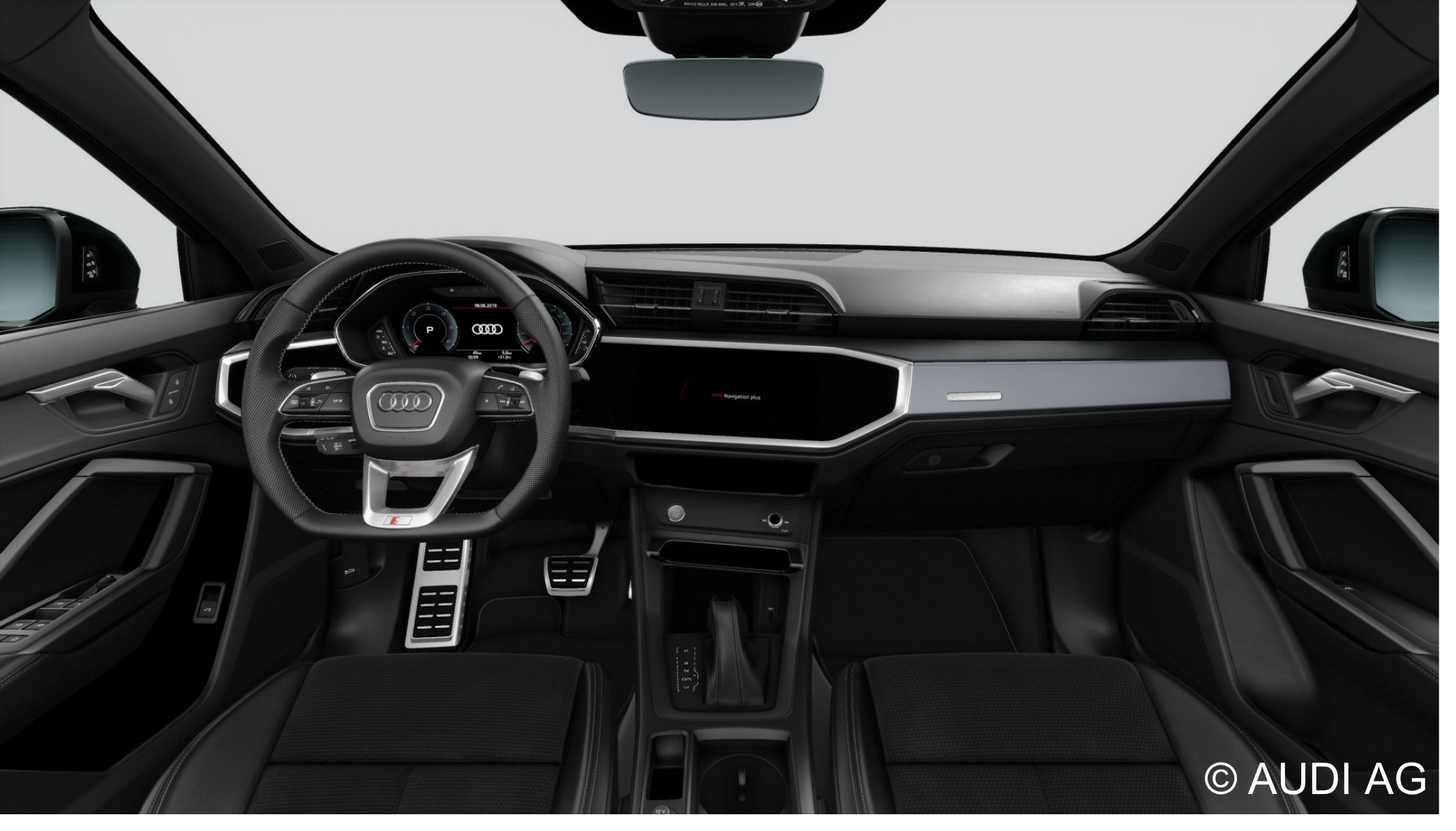}
  \end{minipage}
    \begin{minipage}[b]{0.24\textwidth}
    \includegraphics[width=\textwidth]{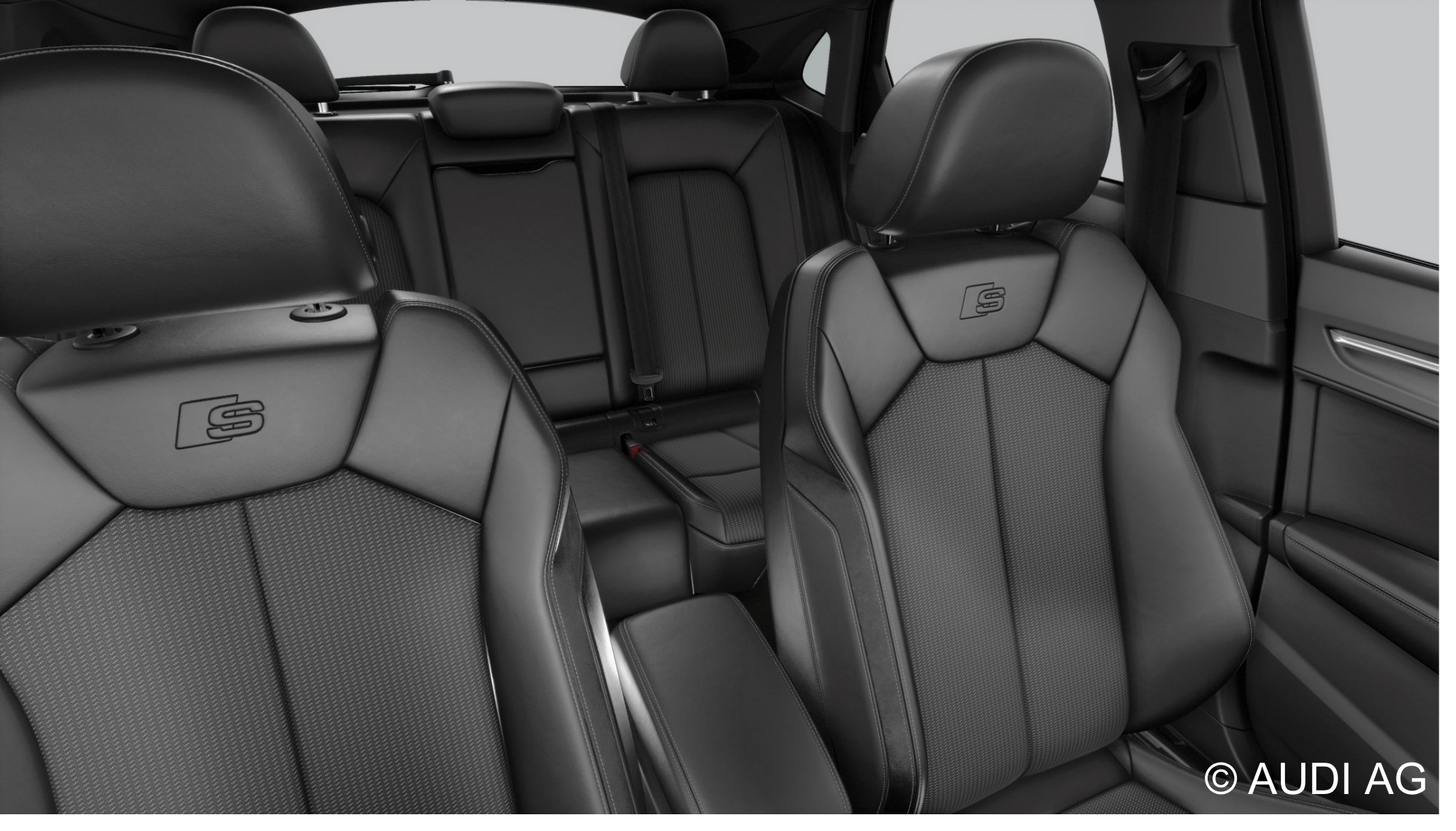}
    \includegraphics[width=\textwidth]{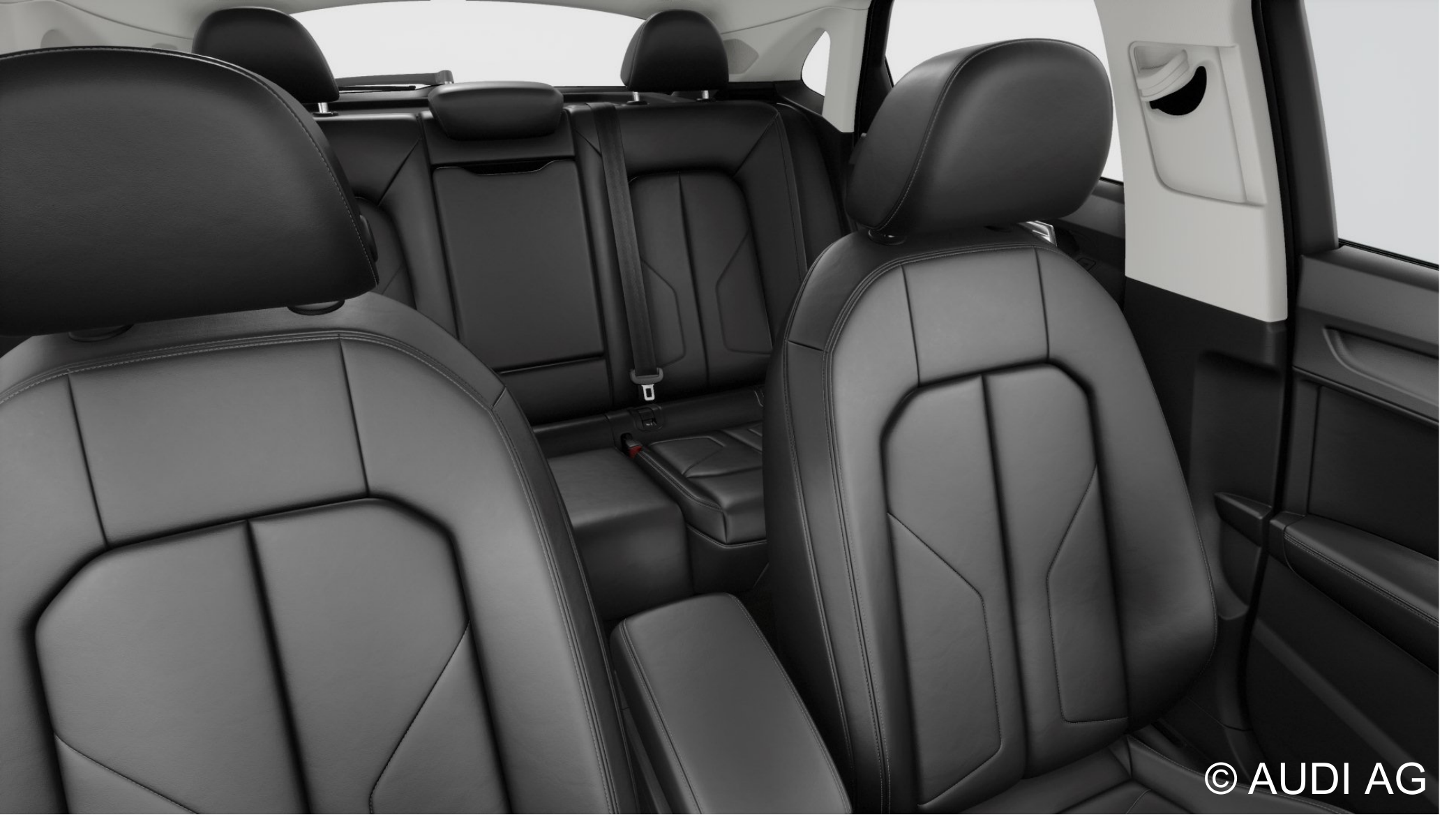}
  \end{minipage}
  \caption{Exemplary Audi Q3 Sportback configurations (from left to right: exterior front view, exterior rear view, interior front view, interior rear view). The upper row displays correct configurations whereas the lower row depicts defective ones. We remove the copyright sign during preprocessing.}
  \label{fig:exemplary_car_images2}
\end{figure*}
Advances in CGI over the last years have paved the way for this technology's applicability to a wide range of use cases in practice.
Many areas encompassing visual experiences, such as 3D animation, visual movie effects, or augmented and virtual reality, have experienced great success due to the increasingly realistic representation possibilities \cite{huang2017rating}.
Today, car manufacturers leverage CGI of their automotive fleet especially in the context of advertisement purposes.
Digital 3D models of all vehicle types including derivatives and vehicle configurations that customers can purchase on the market are created and regularly updated.
Not only does this technology contribute to efficiently scalable advertisement options, e.g., customization to different markets worldwide without the need for real-world photo shoots, but it also offers customers the possibility to compile and experience their preferred car configuration before the purchase decision. 
However, representing all car types including all possible configurations as realistically as possible without any defects while continuously updating them is resource-intense and difficult to maintain by manual means.

In the use case at hand, within the virtual production process of 3D car models at the case company, specific configurations can result in parts of the vehicle not being displayed correctly, e.g., triggered by wiring errors of the individual model components. In these cases, e.g., black areas appear instead.
To readjust them, occurred defects must first be identified, which is currently done in manual random checks based on tacit knowledge. 

\Cref{fig:exemplary_car_images2} depicts typical errors that can occur in the model.
The upper row displays correct model configurations from four different camera perspectives, while the lower row presents respective defects.
In the exterior front view, the wheel arch linings are not displayed, in the exterior rear view the taillights are missing.
Moreover, in the interior front view, among other parts, the upper part of the gear stick is not present, and in the interior rear view the driver's seat has no belt.     

As the manual QA process on-site functions based on randomly chosen configuration and is especially driven by implicit knowledge of which configurations tend to encounter errors, this procedure carries the risk that newly occurring errors remain initially undetected.
However, the goal is to identify as many defects as possible before a car model goes live to avoid sub-optimal customer experiences.  

\section{Artifact Design}
\label{sec:artifactdesign}
In line with the design science research paradigm, we start by elaborating on our design choices \cite{meth2015designing}. First, we are confronted with necessary design requirements (DRs) for the artifact.
\begin{table}[htbp]
\centering
\caption{Overview of workshop participants.}
\label{tab:workshop}
\begin{tabular}{|l|l|l|}
\hline \bf Participant & \bf Focus & \bf Role\\ \hline
Alpha & QA & Senior Manager \\
Beta & QA & 3D Specialist  \\
Gamma & QA/ML & Researcher \\
Delta & ML & ML Researcher \\
Epsilon & ML & ML Researcher \\
\hline
\end{tabular}
\end{table}
\begin{table*}[!htbp]
\centering
\caption{Overview of design knowledge.}
\label{tab:designoverview}
\resizebox{\textwidth}{!}{%
\begin{tabular}{|l|l|l|l|c|}
\hline
\multicolumn{1}{|c|}{\textbf{Design Requirement}}                                                                                                       & \multicolumn{1}{c|}{\textbf{\begin{tabular}[c]{@{}c@{}}Interview \\ Source\end{tabular}}} & \multicolumn{1}{c|}{\textbf{\begin{tabular}[c]{@{}c@{}}Literature \\ Source\end{tabular}}} & \multicolumn{1}{c|}{\textbf{Design Principle}}                                                                                                                                                                       & \textbf{\begin{tabular}[c]{@{}c@{}}Design \\ Kernel Theory\end{tabular}}                       \\ \hline
\begin{tabular}[c]{@{}l@{}}\textbf{DR1}: The artifact should require\\ significantly fewer images than \\ a conventional machine learning\\ approach\end{tabular} & \begin{tabular}[c]{@{}l@{}}Alpha, Beta, \\ Gamma, Delta, \\ Epsilon\end{tabular}            & \cite{schohn2000less, khan2019}                                                                                         & \begin{tabular}[c]{@{}l@{}}\textbf{DP1}: Provide the artifact with \\ active machine learning \\ capabilities to identify instances \\ with the highest contribution to \\ the training period\end{tabular}                   & \multirow{3}{*}{\begin{tabular}[c]{@{}c@{}}Active\\ machine\\ learning\\ \cite{settles2009active}\end{tabular}} \\ \cline{1-4}
\begin{tabular}[c]{@{}l@{}}\textbf{DR2}: The artifact's performance \\ should be equal to \\ a conventional machine \\ learning approach\end{tabular}             & \begin{tabular}[c]{@{}l@{}}Alpha, Beta, \\ Gamma, Delta, \\ Epsilon\end{tabular}            & \cite{pereira2019empirical, baram2004online}                                                                                         & \begin{tabular}[c]{@{}l@{}}\textbf{DP2}: Provide the artifact with \\ active machine learning \\ capabilities to increase the \\ learning curve significantly \end{tabular} &                                                                                                \\ \cline{1-4}
\begin{tabular}[c]{@{}l@{}}\textbf{DR3}: The artifact should\\ ensure no or only a fraction\\ of erroneous images are missed\end{tabular}                        & \begin{tabular}[c]{@{}l@{}}Alpha, Beta, \\ Gamma, Delta, \\ Epsilon\end{tabular}            & \cite{cormack2016scalability, abualsaud2018system}                                                                                         & \begin{tabular}[c]{@{}l@{}}\textbf{DP3}: Choose a performance \\ metric with a focus on recall;\\ $F_2$ metric\end{tabular}                                                                                       &                                                                                                \\ \hline
\begin{tabular}[c]{@{}l@{}}\textbf{DR4}: Identified images \\ that contain errors should \\ be forwarded to QA managers\end{tabular}                             & Alpha, Beta, Gamma                                                                        & \cite{emmanouilidis2019enabling, vossing2019designing}                                                                                         & \begin{tabular}[c]{@{}l@{}}\textbf{DP4}: Provide the artifact \\ with a forwarding mechanism \\ to allow QA managers \\ direct feedback\end{tabular}                                                                          & \begin{tabular}[c]{@{}c@{}}Human-in-the-loop\\ \cite{zanzotto2019human}\end{tabular}                            \\ \hline
\end{tabular}
}
\end{table*}

In the work at hand, we design an ALQA system with the goal to support the identification of defects in car renderings. We conducted an initial focus group workshop \cite{morgan1996focus} with two experts (Alpha, Beta) from our case company as well as three ML researchers (Gamma, Delta, Epsilon). An overview of the roles and foci is depicted in \Cref{tab:workshop}. In the workshop, we discussed the needs and specifications for the solution and decided on the most important aspects. Based on the results of this workshop as well as related literature, we now derive a set of DRs for the proposed artifact. 

We discovered that the case company needs the artifact to require significantly fewer images than conventional ML approaches like supervised ML (DR1). While they have strategically decided to introduce ML techniques to support QA in general, they are hesitant because typically large data sets are required for training \cite{chen2014big}. The number of images rendered, the small rate of errors in the images, and the fact that ground truth images cannot automatically be generated with respective labels at the beginning of the project require that the artifact should need as little labeled input as possible. Furthermore, as the task is currently performed manually by humans with high accuracy rates, it is important that performances are comparable. 
While conventional approaches have shown to reach high performances in similar tasks \cite{zheng2020generic}, it remains unclear how other approaches perform. As a consequence, the performance of conventional approaches must be at least reached (DR2) and no or only a small fraction of images should be missed (DR3). Finally, those images that are identified as erroneous should be forwarded to a QA manager for the final decision (DR4).

In order to address the requirements, we select relevant theories from the body of knowledge, which are then incorporated into design principles (DPs). We address DR1--3 with the kernel theory of \textit{active (machine) learning} (AL) \cite{settles2009active}. AL is a special case of ML in which the respective algorithm can interactively query an information source (e.g., a human) to label new data points with the desired outputs. As an AL algorithm, we select DEAL \cite{9356322}. We provide more details on AL, possible implementation options, and reasons for the choice of DEAL in the upcoming \Cref{sec:rigorcycle}. By utilizing AL, we aim to select those instances within the data set which have the highest contribution to the training process (DP1). This should then result in a significant increase in the artifact's ability to learn faster (DP2). To evaluate the approach, a performance metric that minimizes false negatives is desired as our goal is to avoid defects remaining undetected. Thus, we choose the $F_2$ metric (see \Cref{eq:fbeta}) as it puts a focus on recall without completely ignoring precision (DP3).
%
%
As DR4 is rather a procedural requirement than a technical one, we inform our design by utilizing the theory of \textit{human-in-the-loop} \cite{zanzotto2019human}. Human-in-the-loop is a theory that prescribes human interaction within decision processes that are supported by ML. By providing the deployed artifact with a forwarding mechanism, we allow QA managers direct feedback of errors so they can investigate root causes (DP4).

The resulting DPs, a summary of the DRs, as well as the related theories are shown in \Cref{tab:designoverview}. These requirements are of general nature and are supported by both interviews and literature.

\section{Rigor Cycle: Related Work and Research Gap}
\label{sec:rigorcycle}
In this section, we ensure rigorous research by elaborating on related work. To that end, we specifically focus on QA in CGI as well as AL.
\subsection{Quality Assurance in Computer Generated Imagery}
The utilization of deep learning for computer vision tasks has shown remarkable results, especially in the area of image recognition \cite{he2016deep}. Naturally, besides other application cases, the technology found its application soon within the area of QA tasks \cite{stadelmann2018deep}. This field of research is often also called \textit{image quality assessment} \cite{bosse2017deep}. Computer vision in general and deep learning in specific are already utilized for a large variety of QA procedures. Examples include the control of printing quality \cite{villalba2019deep},
quality estimation of porcelain \cite{onita2018quality}, or wear characteristics of machining tools \cite{walk2020towards}.

Apart from traditional QA within physical industrial processes, quality control of CGI is an especially interesting area for our work---as we deal with digital car renderings. Interestingly, the topic of QA in CGI is little explored. While some works do generate images in a synthetic way as a basis for training models \cite{mahapatra2018efficient}, the focused detection of errors in CGI is rare. The closest work relevant in this regard is the area of video game streaming \cite{zadtootaghaj2020quality}. As cloud gaming solutions like Google Stadia, Playstation Now or Nvidia Geforce Now \cite{di2020network} allow the streaming of video games via the internet, a niche group of researchers aims to detect quality issues in the generated video streams \cite{utke2020towards, zadtootaghaj2018nr, barman2018evaluation}. However, to the best of our knowledge, no work previously explored QA for CGI within an industrial context.
\subsection{Active Machine Learning}
Active (machine) learning (AL) has been intensively researched over the past decades.
%
%
In general, AL methods can be divided into \textit{generative} (e.g., \cite{mayer2020adversarial,mahapatra2018efficient}) and \textit{pool-based} (e.g., \cite{sener2017active,kirsch2019batchbald,ash2019deep,wang2016cost,yoo2019learning,9356322,beluch2018power,gal2017deep}) approaches. 

\textit{Generative} methods employ generative adversarial networks (GANs) to create informative samples which are added to the training set \cite{mayer2020adversarial,mahapatra2018efficient}.
%
%

\textit{Pool-based} approaches utilize various acquisition strategies to select the most informative data instances.
The literature further distinguishes \textit{diversity-} (e.g., \emph{core-set} \cite{sener2017active}) and \textit{uncertainty-based} approaches, as well as combinations of both (e.g., \emph{Batch BALD} \cite{kirsch2019batchbald} or \emph{BADGE} \cite{ash2019deep}).
For our work, uncertainty-based techniques are most relevant.
Here, the underlying assumption is that the more uncertain a model is with respect to a prediction, the more informative the associated data point has to be.
Uncertainty-based AL approaches can be further divided into \textit{ensemble-based} (e.g., \emph{Deep Ensemble} \cite{beluch2018power} or \emph{MC-Dropout} \cite{gal2017deep}) or \textit{non-ensemble-based} (e.g., \emph{Minimal Margin} \cite{wang2016cost}, \emph{Learning Loss} \cite{yoo2019learning}, or DEAL \cite{9356322}) methods.
%
Generally, deriving uncertainty estimates via the softmax function carries the risk that a model may be uncertain in its predictions despite high softmax probabilities \cite{gal2016uncertainty}.

%
Approaches addressing this drawback, e.g., generate multiple predictions per instance, which are subsequently aggregated, by using the technique of \emph{MC-Dropout} \cite{gal2017deep} or by employing a model ensemble (\emph{Deep Ensemble} \cite{beluch2018power}). Another approach is to add a separate loss prediction model to the network \cite{yoo2019learning}.
%
While the first two approaches are time-consuming regarding the acquisition of new data instances, the latter requires implementing an extra model.

In this work, we employ DEAL \cite{9356322} as it allows the selection of new data through uncertainty estimates generated by only one forward pass of each instance through the network. These uncertainty estimates are derived by placing a Dirichlet distribution on the class probabilities. Its parameters are set by the output of the neural network \cite{sensoy2018evidential}.

\section{Artifact Evaluation}
\label{sec:artifactevaluation}
In this section, we first describe the utilized data (\Cref{sec:dataset}), implement the previously described artifact design of the ALQA system (\Cref{sec:experimentalsetup}), and evaluate it accordingly with a common AL procedure \cite{settles2009active} (\Cref{sec:perfeval}). 
Here, the focus of the evaluation is on the AL component of the ALQA system, i.e., whether a performance comparable to a conventional ML component can be achieved with fewer data instances to-be-labeled and whether its classification performance is sufficient for practical use. The human-in-the-loop component is not evaluated because the only source of error results from the AL component.  
\subsection{Data Set}
\label{sec:dataset}
During the initial focus group workshop, we chose the 3D model of the Audi Q3 Sportback for the evaluation of our artifact as we were able to draw on a model status that had (at the time of the conducted project) not yet undergone the QA process on site. Thus, we can assess during the evaluation whether our artifact is capable of detecting the defects that have been identified in the manual QA process afterward. 
Since ML approaches applied directly to 3D data, e.g., 3D point clouds, are still in their infancy \cite{sager2021labelcloud}, we reduce the problem complexity to 2D images \cite{guodeep}. However, the artifact requires analyzing the largest possible area visible to customers during the QA process in order to ensure no visible key areas are missed. Thus, we select the four camera perspectives introduced in \Cref{sec:relevancecycle} to fulfill this requirement (see \Cref{fig:exemplary_car_images2}).

As our computational resources are limited, we randomly select a subset of 4,000 uniquely identifiable vehicle configurations from a pool of customer orders. As we employ an untested car model, it inherently contains defective configurations. Furthermore, in order to represent a large diversity of theoretically manifestable defects, we randomly specify parts of the vehicle not being displayed correctly by the rendering engine in half of the subset. Subsequently, we render the entire pool of car configurations for each camera perspective. 
As the model inherently contains defective configurations and the procedure of sampling additional defects might affect parts invisible to the customer, we cannot automatically generate ground truth labels for the rendered data set. Consequently, each image must be subjected to manual labeling. Thus, the entire data set was labeled within the focus group over a period of several weeks. Ambiguous instances were discarded. Finally, for each camera perspective, we assign a fixed number of 2,000 images to the training set and allocate the remaining images to the test set after reserving 10\% of the data per camera perspective for the validation set. 
\subsection{Experimental Setup}\label{sec:experimentalsetup}
\begin{figure*}[!h] 
\begin{multicols}{2} 
    \centering
    \includegraphics[width=0.878\linewidth]{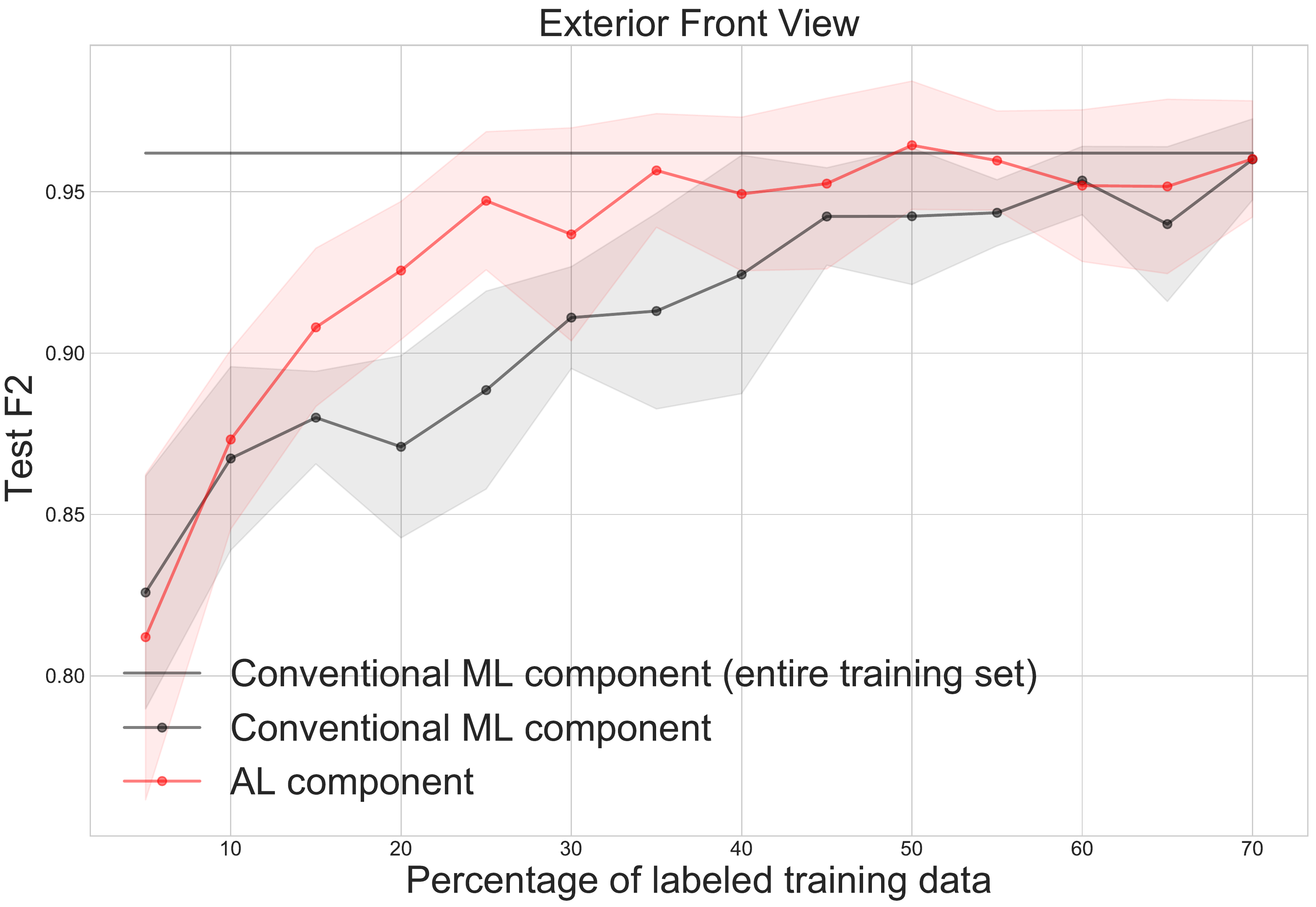}\par 
    \includegraphics[width=0.878\linewidth]{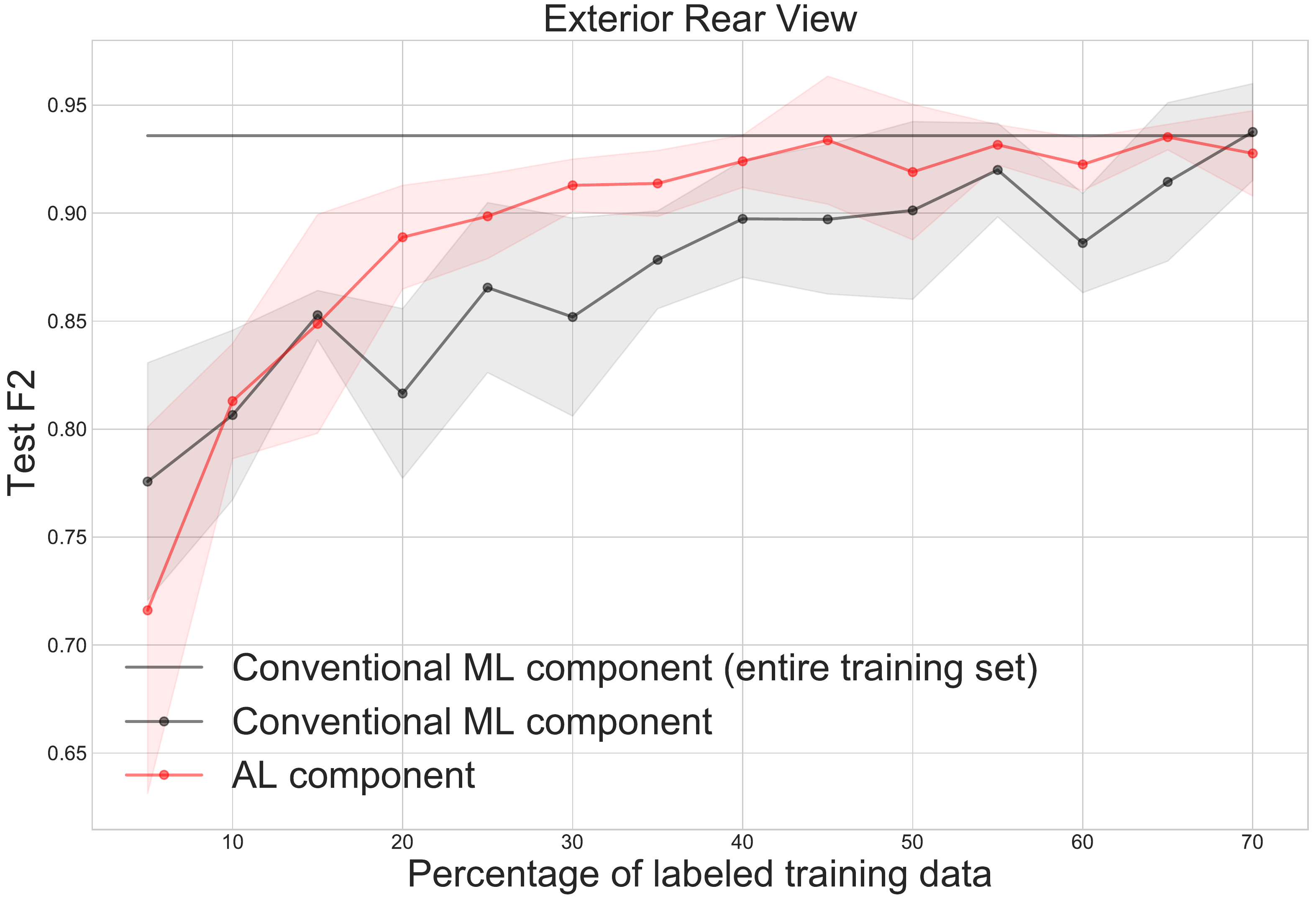}\par 
    \end{multicols}
\begin{multicols}{2}
    \centering
    \includegraphics[width=0.878\linewidth]{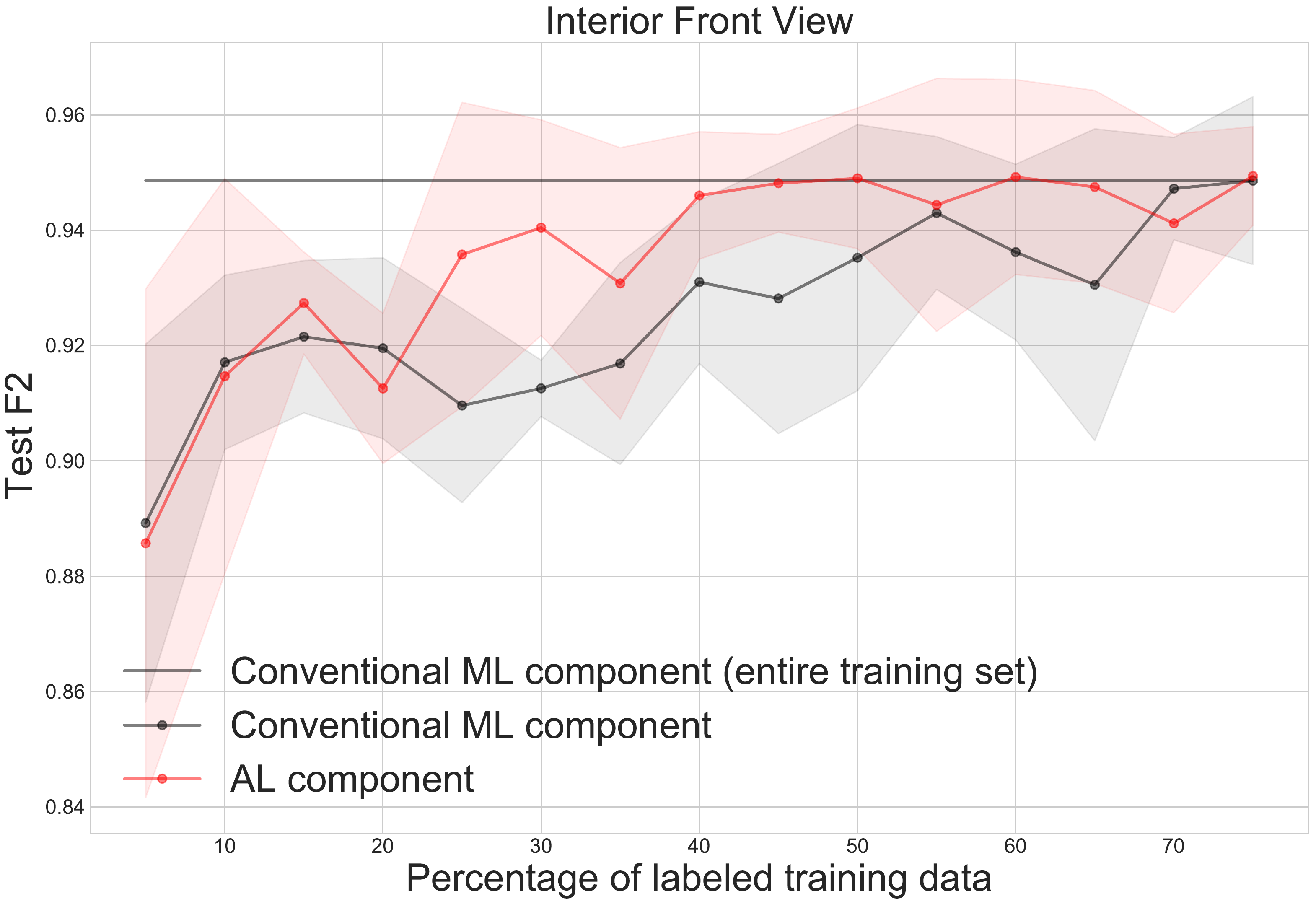}\par
    \includegraphics[width=0.878\linewidth]{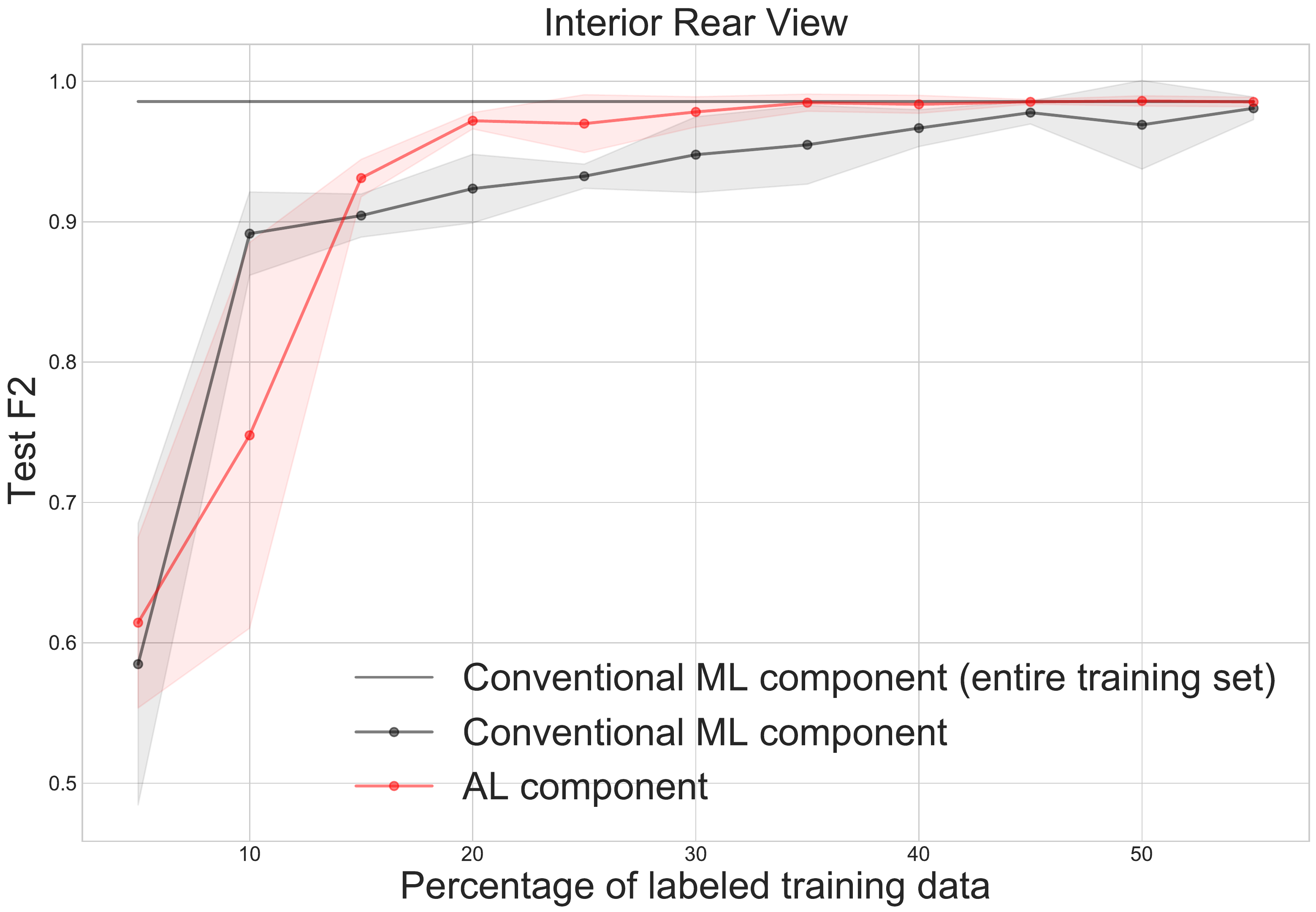}\par
\end{multicols}
\caption{$F_2$ metric over the percentage of acquired labeled training data for exterior front view, exterior rear view, interior front view, and interior rear view. We benchmark the performance of our AL component against a conventional ML component that acquires new data uniformly at random. The solid horizontal line represents the conventional ML component trained with all labeled training data. Shaded regions display standard deviations.}
\label{fig:active_learning}
\end{figure*}
To demonstrate the effectiveness and efficiency of our approach, we evaluate it according to the following AL scheme: Initially, a model is trained on a small labeled data set. In each AL round, new data instances are chosen by an acquisition function, labeled by an expert, and added to the pool of labeled data. This procedure is repeated until a predefined termination criterion is met \cite{settles2009active}. 

As a classification model, we choose a convolutional neural network (CNN) using the ResNet-18 architecture \cite{he2016deep} with the modifications and acquisition function as described by \citet{9356322}. All images are compressed to a resolution of 128$\times$128 pixels. Initially, for each camera perspective, we provide the model with a small labeled data set consisting of 100 randomly sampled data instances. Hereupon, the model is trained from scratch for 100 epochs while monitoring the validation performance metric development. Early stopping is applied once the performance metric of the validation set does not increase within 20 epochs. 
We choose a training batch size of 50, a learning rate of $5\cdot10^{-4}$, and Adam \cite{kingma2014adam} as optimizer. In each AL round, a new batch with 100 images is selected and added to the training data pool. Afterward, the model is re-trained from scratch again. 

We benchmark our approach against a conventional ML component that is trained using the entire available training set and serves as an upper performance bound. Moreover, to compare the learning progress to our approach, the conventional ML component is additionally tested using the aforementioned AL evaluation. However, new batches of images are acquired uniformly at random. We terminate the AL procedure for the conventional ML component once its performance differs only marginally from its performance on the complete training data set. 

For each camera perspective, we apply the resulting number of AL rounds equally to our AL component. This allows us to determine if the AL component is able to reach the final performance level of the ML component with fewer labeled images.

We repeat each experiment five times and report the $F_2$ metric (see \Cref{eq:fbeta}) including standard deviation. As the artifact's purpose is the identification of defective configurations, minimizing false negatives becomes more important than minimizing false positives. Since the $F_2$ metric places more focus on recall, without neglecting precision \cite{goutte2005probabilistic}, it is suitable for our case.
\begin{equation}
\label{eq:fbeta}
F_2 = 5 \cdot \frac{precision \cdot recall}{4 \cdot precision + recall}
\end{equation}
\subsection{Performance Evaluation}
\label{sec:perfeval}
We conduct the experimental evaluation as described in \Cref{sec:experimentalsetup}. The results for all four camera perspectives as a function of the number of training instances are depicted in \Cref{fig:active_learning}. The horizontal lines represent the performance of a conventional ML component trained on the entire training data set. For the exterior views, $F_2$ scores of $96.20\%$ (front) and $93.58\%$ (rear) can be achieved. The interior views result in $F_2$ scores of $94.86\%$ (front) and $98.58\%$ (rear), respectively. 
While repeatedly adding new data instances after each AL round to the initial training data pool, we see that the AL component identifies new data instances contributing to an overall steeper learning curve in comparison to the conventional ML component. For all camera perspectives, the AL component reaches the upper bound in terms of $F_2$ score using substantially fewer training data instances. Thus, DR1, DR2, and DR3 are addressed successfully. 

The savings can be extracted from \Cref{fig:active_learning} via the distance of the touchpoints of the red and black learning curves with the horizontal line. Specifically, a saving of 400 images can be achieved for the exterior front view, 500 for the exterior rear view, 600 for the interior front view, and 400 for the interior rear view. 

To substantiate these findings, we apply a paired t-test assessing if the respective performance levels differ significantly between the AL and ML component. Its assumptions are validated beforehand. The criterion of normally distributed differences of the pairs is validated based on \citet{shapiro1965analysis}. We note that the null hypothesis that a sample comes from a normal distribution cannot be rejected for the exterior front view $(W_s = 0.9106, p = 0.1606)$, exterior rear view $(W_s = 0.9244, p = 0.2540)$, and interior front view $(W_s = 0.9258, p = 0.2361)$. Performing the paired t-test results in the null hypothesis to be rejected 
for the exterior front view $(t =3.6355 , p = 0.0030)$, exterior rear view $(t = 2.3617, p = 0.0345)$, and interior front view $(t = 3.0639, p = 0.0084)$. As the null hypothesis for the Shapiro-Wilk test is to be rejected 
for the interior rear view $(W_s = 0.6671, p = 0.0002)$, following literature, we perform a Wilcoxon signed-rank test \cite{wilcoxon1992individual}. Here, the null hypothesis that two related paired samples come from the same distribution can be rejected $(W_w = 11.0, p = 0.0537)$. 

Finally, we briefly discuss the observed results. A possible reason explaining both the performances of the conventional ML and the AL component might be the image quality. Unlike most real-world use cases, the rendering engine provides the ability to render any vehicle configuration with the same quality and camera angle. Another aspect we want to highlight is the quality of the labeled instances. Potential errors in the labeling process by human annotators can negatively influence the overall model performance. This might explain the upper performance bounds of the models trained with the entire available training set and the observed decline in the performance of the AL component between individual AL rounds (see \Cref{fig:active_learning}).

\section{Discussion}
\label{sec:discussion}
With the results at hand, we discuss a potential economic impact. We analyze possible time savings resulting from employing the ALQA system in comparison to a conventional ML-based approach and contemplate its embedding in the overall QA process.
Even though more and more data is becoming available nowadays for catalyzing the usage of supervised ML algorithms \cite{chen2014big}, necessary prerequisites, e.g., available labels are often not met in practice \cite{baier2019challenges}. Therefore, we intend to demonstrate with the following calculation-based example that AL-based approaches have the potential to mitigate these issues. 
However, our calculations are based on multiple assumptions, e.g., similar performances across all models and the same labeling quality by respective annotators for all other car types of the fleet. Thus, they should be interpreted as indicative to illustrate the potential of ALQA and similar approaches. 
In the case of the regarded model (Audi Q3 Sportback), it was not possible to collect data instances including ground truth labels. Thus, manual labeling was inevitable and was conducted within the focus group over several weeks (see \Cref{sec:dataset}). 
For a randomly selected subset of images, we measured the time it took each focus group member to label each instance resulting in an average labeling time of 31 seconds (standard deviation 23 seconds) per image.   
By incorporating the AL component in the artifact, an average of 475 fewer images had to be labeled per camera perspective compared to a conventional ML component. The total savings of 1,900 images are equivalent to an average reduction of $35\%$.
Taking into account the average labeling time per image and a working day of 8 hours \cite{workinghours}, the labeling effort is reduced by approximately two full days per car model.
Now, let us imagine extending the ALQA system to the entire fleet of the case company. 
Considering 18 vehicle types \cite{modeloverview} would increase the potential average savings to 36 days, which emphasizes the potential to lower initial barriers for the practical use of such a system. 

Finally, we want to stress that the ALQA system's objective is to complement human capabilities in the QA process, not to replace them \cite{hemmer2021human}. As a large number of configurations can be assessed by the artifact efficiently, identified errors are passed to QA managers for further root cause analysis fulfilling DR4. 
Unblocked capacities taken up by the mere identification of defects can be used more efficiently in tasks not covered by the system, e.g., verifying that dynamic animations work properly.

\section{Conclusion}
\label{sec:conclusion}
Increasingly realistic representation possibilities driven by technological advances in the field of computer-generated imagery have made this technology applicable to a continuously increasing number of industrial use cases. Automotive manufacturers utilize computer-generated imagery for scalable advertisement campaigns and offer customers the possibility to configure a selected car model in online configurators according to their personal preferences. 
However, human-led quality assurance based on visual inspection faces the challenge to keep up with these scalable concepts in terms of efficiency and effectiveness, turning high-volume visual inspection processes into procedural bottlenecks.
Even though deep learning approaches for computer vision tasks \cite{he2016deep}, and in particular for quality assurance purposes \cite{stadelmann2018deep}, have demonstrated promising results, their applicability to computer-generated imagery within an industrial context remains largely unexplored. 
To address this gap, this work proposes an active machine learning-based quality assurance (ALQA) system for identifying visual defects in computer-generated car imagery.
The system is capable of identifying defective configurations with an average $F_2$ metric of $95.81\%$ for the Audi Q3 Sportback while requiring on average $35\%$ fewer data instances to-be-labeled a priori. 

Through the design, implementation, and evaluation of our artifact, we contribute to the body of knowledge in the following ways: 
The proposed artifact can be utilized as a support tool for human-led quality assurance.
Additionally, we demonstrate the artifact's ability to significantly reduce the manual labeling effort while achieving a desired performance level. 
This has the advantage of lowering initial start-up difficulties for companies aiming to leverage the economic potentials of machine learning. 
Moreover, we contribute with generalizable prescriptive design knowledge by providing precise design principles to address the design requirements of ALQA systems.

However, the scope of this paper is subject to certain limitations. One shortcoming is that the evaluation of the proposed system is currently limited to one vehicle model---namely the Audi Q3 Sportback.
Furthermore, the system is to be understood as a supporting tool for quality assurance to increase its efficiency and effectiveness. It does not claim to automate the entire quality assurance process, as it does not cover the verification of dynamic car model functionalities, such as the opening of doors and trunk.

Apart from these limitations, the options for future research are manifold.
An obvious step may be to extend the proposed system to the entire vehicle fleet. %
Moreover, the system currently classifies configurations based on the features extracted from 2D images. A promising field of research to further improve its performance could be to incorporate further information provided by the rendering engine, e.g., camera distances of pixels, to better recognize whether objects that should be present are in fact present.
Lastly, as new configuration options come onto the market at regular intervals, this could result in less accurate classifications over time. Thus, it could be worthwhile to explore the combination of the proposed system with approaches from the field of concept drift.

\section*{Acknowledgments}
\label{sec:acknowledgments}

We would like to thank Audi Business Innovation GmbH and in particular Lorenz Schweiger for facilitating and supporting this research.

\bibliography{references}
\end{document}